%% file: main.tex
\documentclass[letterpaper, 10 pt, conference]{ieeeconf}  
\IEEEoverridecommandlockouts                              

\usepackage[utf8]{inputenc} 
\usepackage{graphics}   
\usepackage{amssymb, amsmath}   

\usepackage{amsthm}     
\usepackage{microtype}  
\usepackage{epstopdf}   
\usepackage{url}    
\usepackage[noadjust]{cite} 
\usepackage{overpic}    
\usepackage[linesnumbered,ruled,vlined]{algorithm2e}    
\usepackage[dvipsnames]{xcolor} 
\usepackage{subfig}
\usepackage{wrapfig}
\usepackage[font=footnotesize,labelfont=bf]{caption}
\usepackage{multirow}   
\usepackage{footnote}   
\usepackage{soul}   
\usepackage{tikz}   
\let\labelindent\relax
\usepackage[inline]{enumitem}
\usetikzlibrary{tikzmark}
\usepackage{svg}

\usepackage[textsize=scriptsize]{todonotes}

\makeatletter
\let\NAT@parse\undefined
\makeatother
\definecolor{darkblue}{rgb}{0.0, 0.0, 0.55}
\usepackage[bookmarks=true, hidelinks]{hyperref}    
\hypersetup{
     colorlinks   = true,
     linkcolor= darkblue,
     citecolor    = darkblue,
     urlcolor=black
}
\theoremstyle{definition}

\theoremstyle{remark}

\DeclareMathOperator*{\argmax}{arg\,max}

\newcommand{\gpn}{\textsc{GPN}\xspace}
\newcommand{\ppn}{\textsc{PPN}\xspace}

\newcommand{\mcts}{\textsc{MCTS}\xspace}
\newcommand{\dqn}{\textsc{DQN}\xspace}
\def\gopg{go-\textsc{PGN}\xspace}

\def\ours{\textsc{MORE}\xspace}

\font\titlefont=ptmb at 14.8pt
\title{\titlefont
Interleaving Monte Carlo Tree Search and Self-Supervised Learning for Object Retrieval in Clutter
}
\author{Baichuan Huang\quad Teng Guo\quad Abdeslam Boularias\quad Jingjin Yu     
\thanks{B. Huang, G. Teng, A. Boularias, and J. Yu are with the Department of 
Computer Science, Rutgers, the State University of New Jersey, Piscataway, NJ, USA. 
Emails: {\tt\small \{baichuan.huang, teng.guo, abdeslam.boularias, jingjin.yu\}@rutgers.edu}.
This work is supported by NSF awards IIS-1734492 and IIS-1846043, IIS-1845888, CCF-1934924 and IIS-2132972.
}%
}

\begin{document}

\maketitle
\thispagestyle{empty}
\pagestyle{empty}

\begin{abstract} 
In this study, working with the task of object retrieval in clutter, we have developed a robot learning framework in which Monte Carlo Tree Search (MCTS) is first applied to enable a Deep Neural Network (DNN) to learn the intricate interactions between a robot arm and a complex scene containing many objects, allowing the DNN to partially clone the behavior of MCTS. In turn, the trained DNN is integrated into MCTS to help guide its search effort. 
We call this approach learning-guided Monte Carlo tree search for Object REtrieval (\ours), which delivers significant computational efficiency gains and added solution optimality.
\ours is a self-supervised robotics framework/pipeline capable of working in the real world that successfully embodies the System 2 $\to$ System 1 learning philosophy proposed by Kahneman, where learned knowledge, used properly, can help greatly speed up a time-consuming decision process over time.
Videos and supplementary material can be found at 
\href{https://github.com/arc-l/more}{\texttt{\textcolor{blue}{https://github.com/arc-l/more}}}. 
\end{abstract}

\section{Introduction}\label{sec:introduction}
Kahneman \cite{kahneman2011thinking} proposed a thought-provoking hypothesis of human intelligence: in solving real-world problems, humans engage fast or ``System 1'' (S1) type of thinking for making split-second decisions, e.g., speech, driving, and so on. For other decision-making processes, e.g., playing chess, a slow or ``System 2'' (S2) approach is taken, where the brain would perform a search over some structured domain for the best actions to take. 
After repeatedly using S2 thinking to solve a given problem, patterns can be distilled over time and burned into S1 to speed up the overall process. In playing chess, for example, good chess players can instinctively identify good candidate moves. First-time or beginner drivers rely heavily on S2 and gradually converge to S1 as they gain more experience. 
This S2$\to$S1 thinking has gained significant attention and has been explored in many directions in machine learning, including attempts at building machines with consciousness \cite{bengio2017consciousness}.
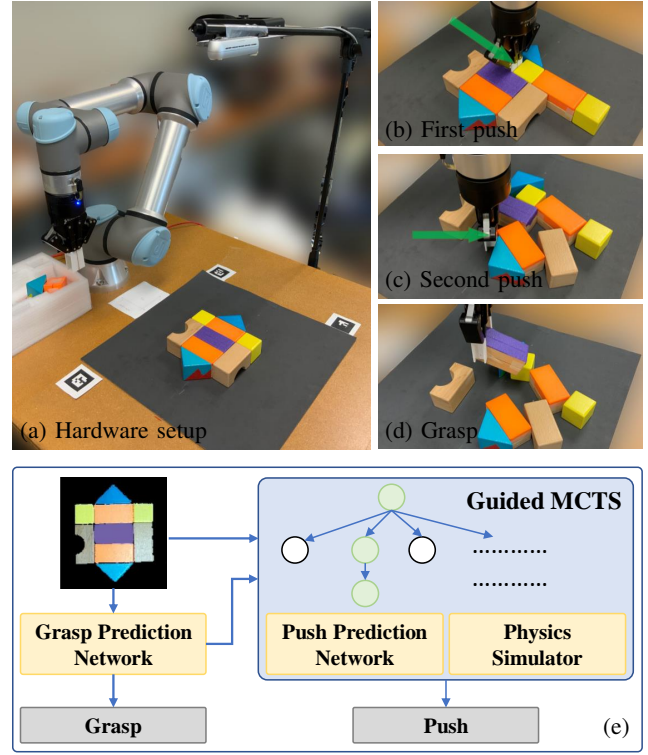
\begin{figure}[t]
\centering
        
        
\vspace{1mm}
\begin{overpic}[width=3.3in,tics=5]{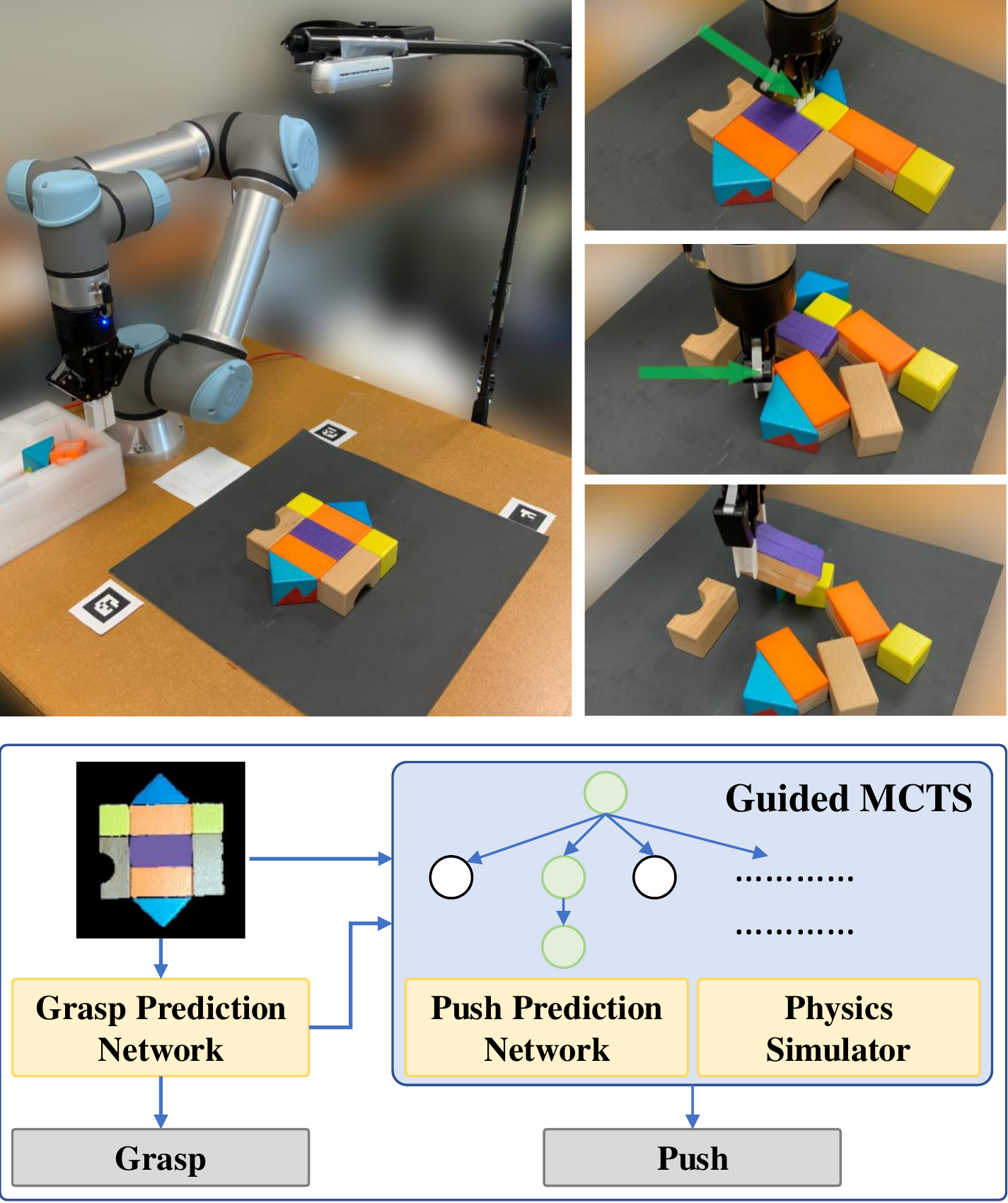}
\put(1,42){{\small (a) Hardware setup}}
\put(49.5,82){{\small (b) First push}}
\put(49.5,62){{\small (c) Second push}}
\put(49.5,42){{{\hypertarget{mylink}{\small (d) Grasp}}}}
\put(78.5,2.5){{\small (e)}}
\end{overpic}
    \caption{\label{fig:intro}
    (a) The hardware setup for object-retrieval-from-clutter includes 
    a Universal Robots UR5e manipulator 
    with a Robotiq 2F-85 two-finger gripper, 
    and an Intel RealSense D455 RGB-D camera. 
    The objects are placed in a square workspace and the target object is masked in \textcolor{RoyalPurple}{purple}. 
    (b)(c) Two push actions (shown with green arrows) are used to enable the grasping of  the target (\textcolor{RoyalPurple}{purple}) object. 
    (d) The target object is successfully grasped and retrieved. 
    (e) The overview of our overall system.
    }
    \vspace{-5mm}
\end{figure}
But, perhaps the most prominent line of work in reinforcement learning \cite{sutton2018reinforcement} that closely aligns with this paradigm is the application of Monte Carlo Tree Search (MCTS) for carrying out self-supervised learning in games \cite{silver2018general,schrittwieser2020mastering}, where an ``understanding'' of a game emerges from a lifelong self-play and is gradually distilled so that it significantly reduces the search effort. Gradually, the overall system learns enough useful information that allows it to play perfect games with much less time and computing resources. 

Inspired by \cite{silver2018general,schrittwieser2020mastering} that show a search-and-learn approach for realizing S2$\to$S1 applies well to game-like settings with relatively well-defined rules, we set out 
to find out whether we could build a similar framework that enables 
real robots to interact with real-world physics and uncertainties to perform physical tasks, somewhat akin to \cite{DBLP:journals/corr/abs-1912-07024}. Specifically, we focus on the task of retrieving an object enclosed in clutter using non-prehensile actions, such as pushing and poking, followed by prehensile two-finger grasping. The goal is to obtain a computationally efficient system and produce high-quality solutions (i.e., using the minimum number of actions).  
%

As pointed out by Valpola \cite{boney2019regularizing}, due to the difficulty in exploring the landscape of the state space of real-world problems, in addition to uncertainty, naive applications of the S2$\to$S1 paradigm often lead to undesirable behavior. 
Non-trivial design as well as engineering efforts are needed to build such S2$\to$S1 systems.
In the object-retrieval-from-clutter setting, the challenge lies in the difficulty of predicting the outcome of push actions, with the tip of the gripper, when many objects are involved.
This is due to discontinuities inherent in object interactions; while a certain pushing action will move a given object, a slightly different direction can miss that same object entirely. 

The main contribution of this work is proving the feasibility of applying the S2$\to$S1 philosophy to build a self-supervised robotic object retrieval system capable of continuously improving its computational efficiency, through \emph{cloning} the behavior of the time-consuming initial MCTS phase. 
Through the careful design and integration of two Deep Neural Networks (DNNs) with MCTS, our proposed self-supervised method, named \textbf{M}onte Carlo tree search and learning for \textbf{O}bject \textbf{RE}trieval (\ours), enables a DNN to learn from the manipulation strategies discovered by MCTS. Then, learned DNNs are fed back to the MCTS process to guide the search. 
\ours significantly reduces MCTS computation load and achieves identical or better outcomes, i.e., retrieving the object using very few strategic push actions. 
In other words, our method ``closes the loop''. This contrasts with \cite{DBLP:journals/corr/abs-1912-07024}, which only learns to replace the rollout function of MCTS. 

\section{Related Work}\label{sec:related}
\input{related-works}


\section{Preliminaries}\label{sec:preliminaries}
The object-retrieval-from-clutter task consists in using a robot manipulator to retrieve a hard-to-reach target object (Fig.~\ref{fig:intro}). 
%
%
Objects are rigid bodies with various shapes, sizes, and colors; the target object is assigned a unique color. 
Similar to~\cite{huang2021visual}, a top-down fixed camera is installed to observe the workspace. 
The camera takes an RGB-D image of the workspace (e.g., the top-left image of Fig.~\ref{fig:intro}), which serves as the only input to our system. 
%

\emph{Pushing} and \emph{grasping} actions are allowed, the execution of each is considered as one \emph{atomic action}.
A grasp action is defined as a top-down overhead grasp motion $a^g=(x, y, \theta)$, corresponding to the gripper's target location and orientation, based on a coordinate system defined over the input image.
%
%
A push action is defined as a quasi-static planar motion $a^p=(x_0, y_0, x_1, y_1)$ where $(x_0, y_0)$ and $(x_1, y_1)$ are the start and the end locations of the gripper tip.
%
The horizontal push distance is fixed and it is $10$cm in our experiments.
Each primitive action is transformed to the real-world coordinates for execution, but all the planning and reasoning are in image coordinates.
%
%
The robotic arm keeps pushing objects until the target object can be grasped or until the target object is pushed outside of the workspace, in which case the task is considered a failure. The problem is to find a policy that maximizes the frequency of successfully grasping the target object, while also minimizing the number of pre-grasp pushing actions.


\section{Methodology}\label{sec:method} 
The \ours framework consists of three components: a Grasp Prediction Network (\gpn), a Monte Carlo Tree Search (MCTS) routine, and a Push Prediction Network (\ppn). 
\gpn is a neural network that predicts the success probabilities of grasp actions. It is trained online similarly to~\cite{huang2021visual}. The success probabilities can be interpreted as immediate rewards. MCTS uses a physics engine as a transition function to simulate long sequences of consecutive push actions that end with a terminal grasp action. Each branch in MCTS is composed of push actions as internal nodes, and a grasp action as a leaf. Grasp actions are evaluated with \gpn, and the returned rewards are back-propagated to evaluate their corresponding branches. The branch with the highest discounted reward, or {\it Q-value}, is selected for execution by the robot. 

While highly effective in finding near-optimal paths, MCTS suffers from a high computation time that makes it impractical. 
To solve this, \ours employs a second neural network, \ppn, to prioritize the action selection in the rollout policy.
The robot starts by relying entirely on MCTS (S2 type of thinking) to solve various instances of the object-retrieval problem. Instead of throwing away the computation performed by MCTS for solving the various instances, we use the computed Q-values as ground-truth to train \ppn. Note that this computation data is free, since it is generated by the simulations performed by MCTS as a byproduct of solving the actual problem. \ppn is a neural network that learns to imitate MCTS and clone its behavior, while avoiding heavy computation and physics simulations by MCTS. As \ppn becomes more accurate in predicting the outcome of MCTS, the robot starts relying on both MCTS and \ppn for action selection. In a nutshell, \ppn is used for orienting the search in MCTS toward more promising push actions that rearranges the scene and renders the target object graspable. After a long experience, \ppn's accuracy in predicting the Q-values of push actions matches that of MCTS, and the robot switches entirely to \ppn to make decisions in a few milliseconds (S1 type of thinking).

%
%

%
%
%
%
%

\subsection{Grasp Prediction Network (\gpn)}
\gpn is a deep neural network based on the model proposed in~\cite{zeng2018learning} and further customized to estimate the expected grasp reward~\cite{huang2021visual}.
We used the pre-trained model from~\cite{huang2021visual}.
with a ResNet-18 FPN~\cite{he2016deep,DBLP:journals/corr/LinDGHHB16} backbone~\cite{huang2020dipn}. For training, only successful grasps are given fixed non-zero rewards. 
%
The Grasp Network takes a single RGB-D image $s_t$ as input and outputs a pixel-wise reward map $R_g(s_t) \in [0, 1]^{H \times W}$ with the same size ($H$ and $W$ are height and width of $s_t$).
%
To enable \gpn to account for gripper orientation, $s_t$ is rotated $16$ times in the range of $[0, 2\pi]$, adding another dimension to the reward map and making it $R_gm(s_t)\in [0, 1]^{H \times W\times k}$ with $k = 16$. Because the goal is to retrieve a specific object, a mask is imposed on the target object using Mask R-CNN~\cite{he2017mask}, effectively truncating the reward map.
If the largest reward from the map $\max_{i,j,\theta}R_{gm}(s_t)[i,j, \theta]$ is larger than some preset threshold, $R_{g*}$, \gpn suggests grasping as the next action to execute. The location $[i,j]$ and rotation $\theta$ of the best grasp is retrieved from the reward map $R_{gm}$.
%

\subsection{Monte-Carlo Tree Search}\label{sec:mcts} 
Monte-Carlo Tree Search (\mcts) \cite{coulom2006efficient} is used in \ours for both decision-making and training \ppn.
%
%
A typical \mcts routine has four steps: selection, expansion, simulation, and back-propagation.
%
In our case, the goal of the search is to find the shortest action sequence; we can stop the search as soon as the best solution is found without exploring the rest.
The search stops in two cases:
\begin{enumerate*}
    \item the number of iterations $n$ exceeds a pre-set budget $N_{max}$, or
    \item the expanded node with state that the target object can be grasped, and all nodes in parent level are expanded.
\end{enumerate*}
%
A node is considered as a leaf if $\max_{i,j,\theta}R_{gm}(s_t)[i,j, \theta] > R_{g*}$ where $R_{gm}(s_t)$ is obtained from \gpn and $R_{g*}$ is a pre-defined high probability. 
%
The maximum depth of the tree is limited to $d$, where $d$ is set to $4$ in our experiments. 

In the selection phase, we find an expandable node starting from the root according to the search policy
\begin{equation}\label{eq:pi}
    \pi_n (s) = \argmax\limits_{a^p}(Q(s, a^p) + C \sqrt{\frac{\ln{N(s)}}{N(s, a^p)}}),
\end{equation}
where $N(s)$ is the number of visits to node (state) $s$ and $N(s, a^p)$ is the number of times push action $a^p$ has been selected at node (state) $s$.
The Q-value is calculated as
\begin{equation}
    Q(s, a^p) = \frac{\sum_{i=1}^{m}{r}_i(s, a^p)}{{\min\{N(n_i), m\}}},
\end{equation}
where $r(s, a^p)$ is the returned long-term reward and $m$ is a pre-set maximum.
Only the best $m$ terms ${r}_i(s, a^p)$ are used to compute the Q-value in the equation above. 
$m$ is set to $10$ when expanding nodes and $1$ when selecting the best solution.
$C$ is the coefficient of the exploration term,  and it is set to $2$ when expanding nodes and $0$ when selecting the best solution.
In the expansion phase, we use a physics simulator to execute the chosen push action $a^p$ at state $s_i$ and predict new state $s_{i+1}$. 
Then, a random policy is used to sample actions to simulate until a grasp is possible or a failure is encountered.
The reward $r$ is predicted by \gpn at a terminal state $s_t$. Reward 
$r$ is set to $1$ if $\max_{i,j,\theta}R_{gm}(s_t)[i,j, \theta] > R_{g*}$, and $0$ otherwise.
One additional term $\delta\max(R_{gm}(s_t))$ is added to $r$, to distinguish between good and bad push actions.
We set $\delta$ to be $0.2$.
In the last step, reward $r$ is propagated back to its parent nodes to update their $Q$-values with a discount factor $\gamma=0.5$. 

\begin{wrapfigure}[9]{r}{1in}
\vspace{-3mm}
  \includegraphics[width=1in]{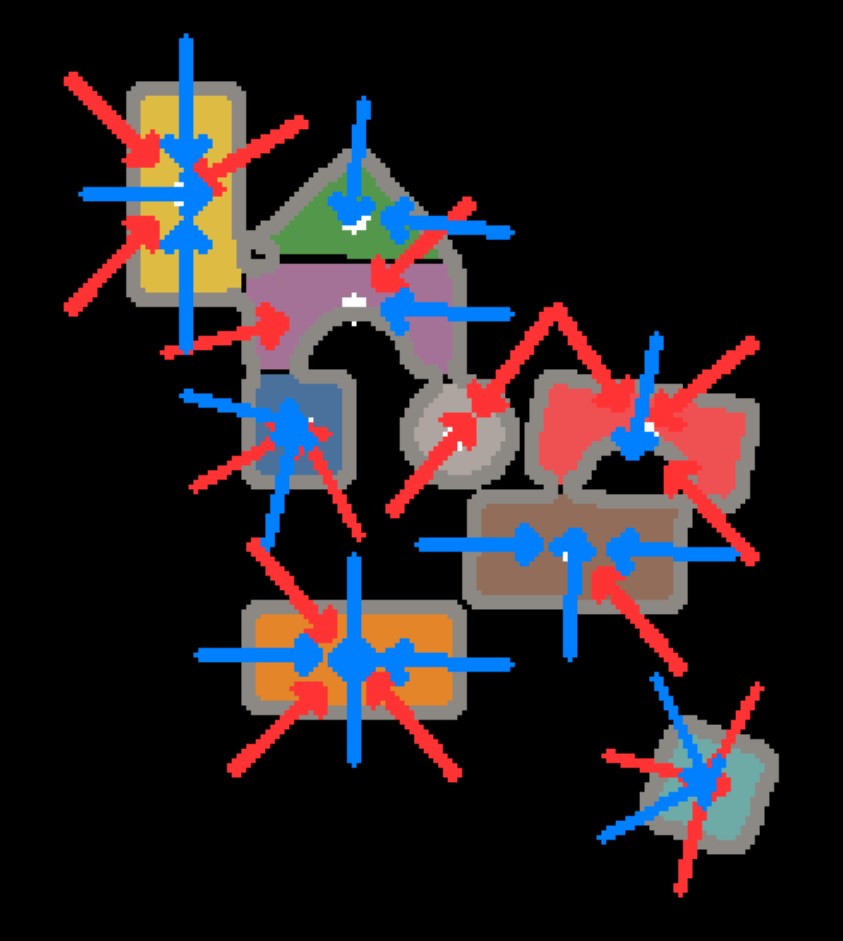}
  \caption{\label{fig:sample}
    Sampled push actions.}
\end{wrapfigure}
As the push action space is enormous even after discretization, we further sample a subset of actions such that all push actions start around the contour of an object and point to the center of the object (Fig.~\ref{fig:sample}).
This action sampling method has been discussed
in~\cite{huang2021visual} and was empirically proven efficient for a similar setup of object retrieval.
%
%

In our implementation, $N_{max}$ is set to $300$ when \mcts is used to collect  data to train \ppn. 
The second and the third conditions for stopping the search are only activated after at least $50$ roll-outs, so that the number of visits to a state is statistically significant and to reduce the variance of \ppn.

\subsection{Push Prediction Network (\ppn)}\label{sec:pushnet}
\begin{figure}[ht!]
    \centering
    \includegraphics[width = .97\linewidth]{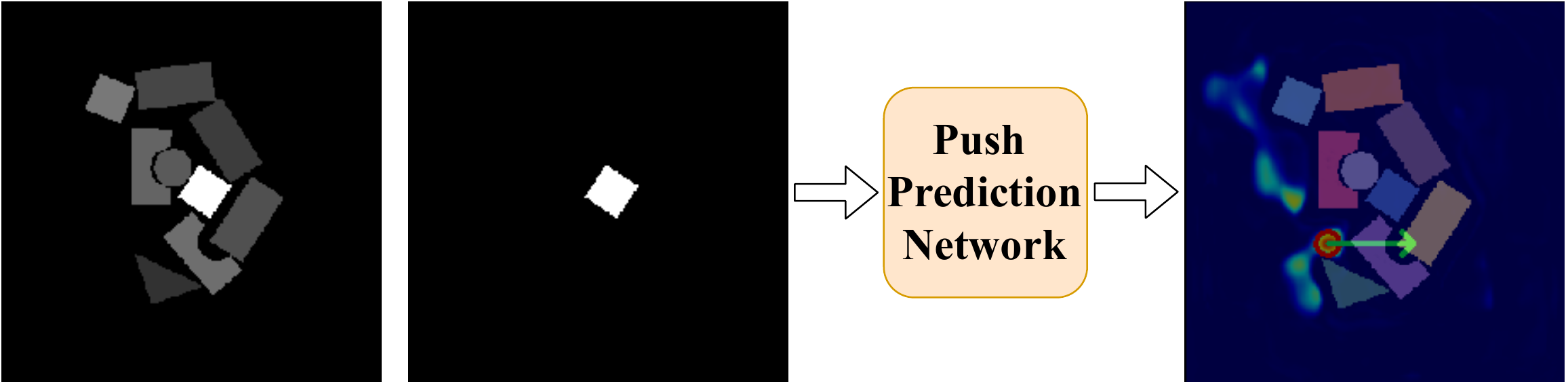}
    \caption{\label{fig:push-in-out}
        The left two figures are the input to \ppn. The first is a segmentation of objects; the second is the mask of the target object. The image on the right is the output from the \ppn. We use Jet colormap to represent the reward value, where the value ranges from red (high) to blue (low). The pixel with the highest Q-value is plotted with a circle and attached with an arrow on the right image, representing pushing action starting at the circle and moving to the right with a distance of $10$cm.
    } 
    \vspace{-2mm}
\end{figure}

As previously mentioned, \ppn learns to imitate MCTS.
\ppn is a deep neural network with ResNet-34 FPN~\cite{he2016deep ,DBLP:journals/corr/LinDGHHB16} as the backbone, where the P2 level of the FPN connects to the head. 
It takes a two-channel input and outputs a single channel pixel-wise push Q-value map, similar to the reward map produced by \gpn.
An example input is shown in Fig.~\ref{fig:push-in-out}, where the first channel is a segmented image of all objects and the second channel is a binary image of the target object.
The output is the image on the right of Fig.~\ref{fig:push-in-out}, where the arrow shows a push action with the highest Q-value.
\ppn estimates the Q-value (discounted rewards) $Q_p(s_t)$ of executing push actions at the corresponding pixel, where the action is assumed to push $10$cm to the right.
$\max(Q_p(s))$ is limited to the range $[0, \eta]$, where $\eta$ is the maximum reward of a terminal state.


When \mcts is used to generate training cases, it builds a tree and saves the transitions for each case: the state (image) $s$, the push action $a^p$, the Q-value $Q(s, a^p)$, and the visited number $N(s, a^p)$.
As such, \ppn is trained in a self-supervised manner.
The input image is rotated based on a push action so that the corresponding push action points to the right.
Because a single action is generated by MCTS (i.e., a $\delta$ signal over the entire input), which is not conducive to training \ppn, we ``expand'' the Q-value over a $3 \times 3$ patch centered around MCTS action but set invalid pushes (e.g., if part of the patch is inside an object) to be zero. 
Now, the label is relatively dense compared to a one-hot pixel, so we can use Smooth L1 loss from Pytorch~\cite{NEURIPS2019_9015} with $\beta$ equals to $0.8$ to regress.
Only gradients on the labeled pixels are used.
Loss weighting is also applied: label values from the \mcts are weighted based on $N(s, a^p)$, label values (zero Q-value) from void push actions are weighted with a small number, $0.001$ for collision and $0.0001$ for pushing thin air.
We observe that \ppn has difficulty learning to create clearance around the target object.
Data augmentation is applied here so that for each training case, we also randomly choose the target object for the \mcts to solve; 
so each arrangement becomes many training cases.
It mitigates over-fitting; given similar visual information, it could learn different strategies, as the target object could be anywhere.

The head model is an FCN with four layers, where the first two layers have a kernel size of 3, the last two 1, and the strides of four layers are all 1.
Batch normalization is used at each layer of the head model except the last.
Bilinear interpolation ($\times 2$) is applied interleaved between the last three layers of the head model to scale up the hidden state to the same size as the input image.
The training process has two stages, one to train the network with a batch size of $8$, learning rate starts at $1\mathrm{e}{-4}$, epoch of $50$.
The learning rate decays with cosine annealing~\cite{loshchilov2016sgdr}, where the maximum number of iterations is set to be the same as the epoch number $50$ and the minimum learning rate is $1\mathrm{e}{-8}$.
The second is a fine-tuning stage; we increase the batch size to $28$ and the learning rate to $1\mathrm{e}{-5}$ with an epoch of $20$.

\subsection{Guided Monte-Carlo Tree Search}\label{sec:guided-search}
With the trained \gpn and \ppn, a guided MCTS is implemented to accelerate the search process, cutting cost from time-consuming expansion and simulation phases.
%
\gpn is again used to determine the terminal state and if so, calculate its estimated reward, as discussed in~\ref{sec:mcts}.
\ppn, trained with data from \mcts, can estimate how much reward can be gained from taking a push action at a certain state.

For this combination of \mcts with \ppn, some additional updates are made (compared to~\ref{sec:mcts}) to incorporate the guidance from \ppn.
The exploration term is removed from the search policy, so $C$ in equation~\eqref{eq:pi} is set to 0.
Similar to~\cite{hamrick2019combining}, we use the estimated reward from \ppn as a prior, so the Q-value is calculated as follows
\begin{equation}
    Q_{guide}(s, a^p) = \frac{\max(Q_p(s)) + \sum_{i=1}^{m}{r}_i(s, a^p)}{N(s, a^p)},
\end{equation}
where $m$ is set to $3$ when expanding nodes and $N(s, a^p)$ is initialized to $1$ for all state-action pairs.
Instead of computing an average as standard MCTS, only best $m$ of $Q_p$ are considered, this is due to the number of rollout is small, a good action could be averaged out.
To select the best action as the next step solution, the Q-value is calculated without the denominator
\begin{equation}
    Q_{best}(s, a^p) = \max(Q_p(s)) + \max({r}_i(s, a^p)),
\end{equation}
where only the best explored solution is considered.

\begin{figure}[ht!]
    \centering
    \includegraphics[width = .97\linewidth]{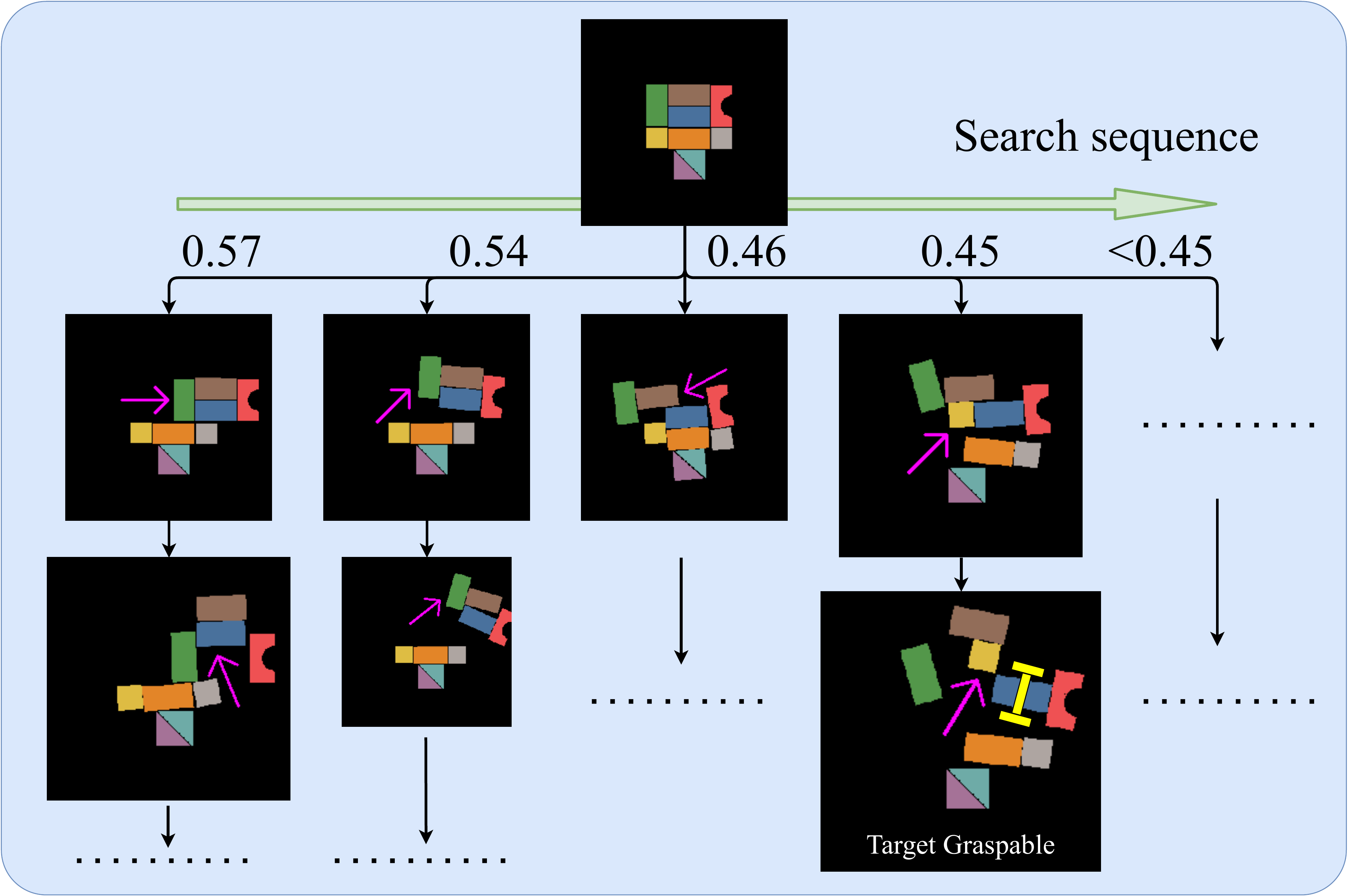}
    \vspace{0mm}
    \caption{\label{fig:guided-mcts}
        An example of the guided MCTS with a budget of 10 iterations. State with larger image have higher estimated Q-values. All expanded nodes are plotted. The numbers in the first levels represent the estimated Q-value returned by \ppn for corresponding push action. These values, together with the reward returned from simulation, guide the tree search.
    } 
    \vspace{-2mm}
\end{figure}

The push action space of the guided MCTS is limited to a subset (like Fig.~\ref{fig:sample}) so that the estimated reward from \ppn is more accurate and the branching factor of the tree is of a reasonable size.
To make the selection mimic the training data, we rotate the image for each sampled push action such that the push action in the rotated image is always pointing to the right.
Then, we only use the estimated Q-value at the corresponding pixel (push action) of the output Q-value map.
%
An example of guided \mcts is given in Fig.~\ref{fig:guided-mcts}.
The expansion of the tree is prioritized by \ppn, where the push action with higher Q-value is sampled earlier, and the rollout policy is also prioritized. 
The maximum depth of the tree is limited to $3$ instead of $4$ as used in the earlier version of MCTS for collecting data to train \ppn.

\vspace{-1mm}

%


\section{Experimental Evaluation}\label{sec:evaluation}
We evaluated the proposed technique both in simulation (PyBullet~\cite{coumans2021}) and on adversarial test cases on a UR5e robot with a Robotiq 2F-85 gripper using real objects.
The robot, workspace, objects, and camera are the same in simulation and real-world experiments, so that we can seamlessly transfer from simulation to the real setup.
The workspace is limited to a square with a side length of $0.448$m; it is discretized as a grid of $224 \times 224$ cells during the image processing step.
The friction of objects and table cannot be accurately measured; nevertheless, high-fidelity physical properties do not seem to be needed for this particular application.
The results demonstrate that the proposed method significantly outperforms MCTS~\cite{huang2021visual} in terms of time efficiency while returning plans of equal quality. The plans returned by the proposed technique contain fewer actions and yield higher success rates than those returned by the purely learning-based solution presented in~\cite{xu2021efficient}. 
%
Training and evaluation are completed on a machine with an Intel i7-9700K CPU and an Nvidia GeForce RTX 2080 Ti.

\subsection{Simulation experiments}
\begin{figure}[ht!]
    \centering
    \includegraphics[width = \linewidth]{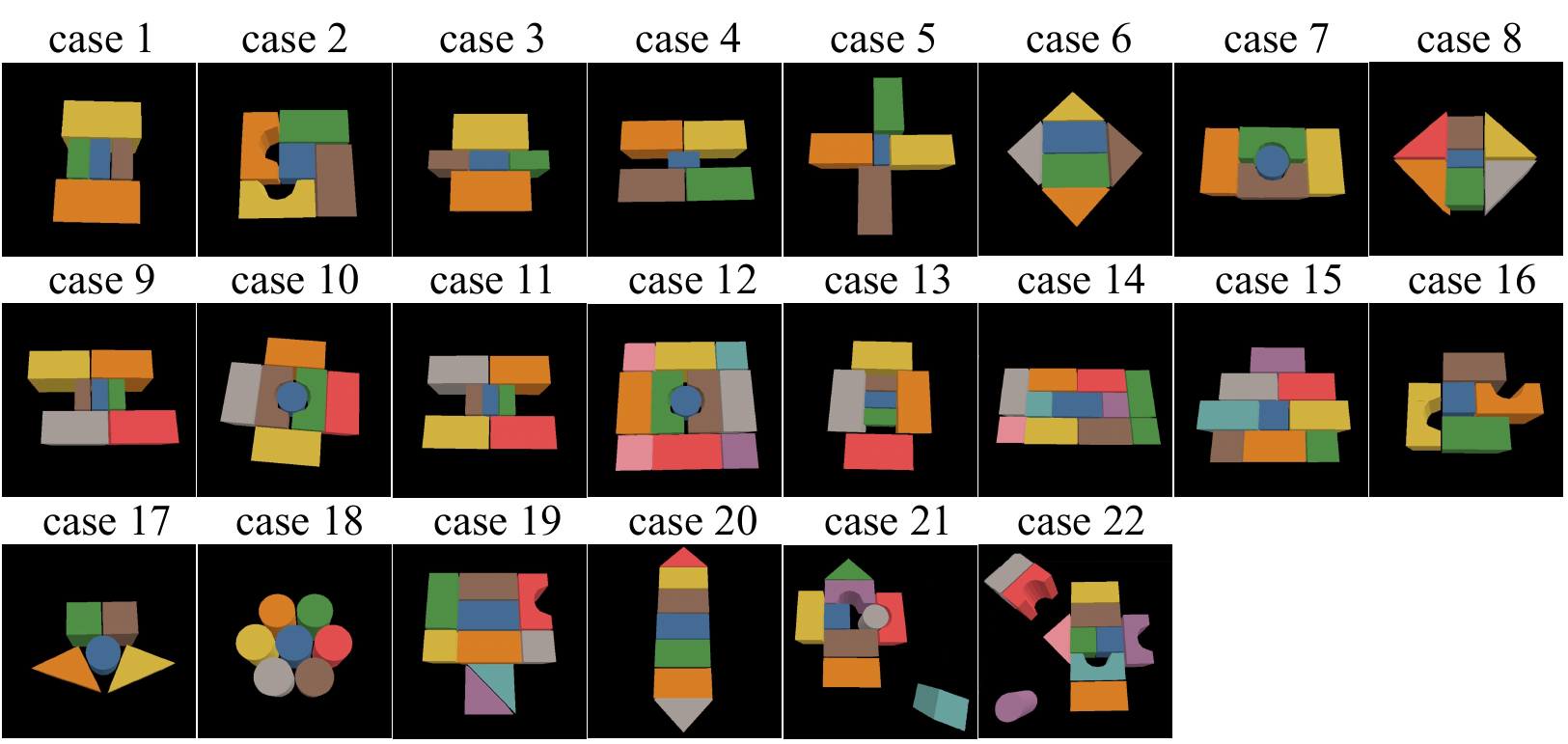}
    \vspace{-3mm}
    \caption{\label{fig:testcases}
        22 cases~\cite{huang2021visual} used in simulation experiments, where the target object is masked in \textcolor{RoyalBlue}{blue} at the center. 
    }
    \vspace{-2mm}
\end{figure}

\noindent \textbf{Tasks.} 
Given an arrangement of heterogeneous and tightly packed objects, a target object is to be retrieved using push and grasp actions from a two-finger gripper.
In simulation, we benchmark on 22 adversarial test cases from~\cite{huang2021visual} (Fig.~\ref{fig:testcases}) and 10 from~\cite{zeng2018learning, xu2021efficient}.
Here ``adversarial'' means that at least one push action has to be executed for a grasp action to be feasible (insert gripper without collision).
Random cases, which are too easy~\cite{huang2020dipn, huang2021visual}, are not discussed here.

\noindent \textbf{Metrics.} 
We use four metrics:
    1) the number of actions used to retrieve the target object,
    2) the total time used for retrieving the target object, which includes both planning time and execution time for simulation results,
    3) the completion rate, failures occur when the target object is pushed out of the workspace, and 
    4), the grasp success rate, which is the number of successful grasps divided by the total number of grasping attempts.
The number of re-arrangement actions that are needed to make the target object graspable and time are the two main metrics.
The completion and grasp success rates are also reported but are not the main focus as they are often close to $100\%$.

\noindent \textbf{Baseline Methods.} We compare with three methods:
\begin{enumerate*}
    \item A self-supervised reinforcement learning method denoted as \gopg~\cite{xu2021efficient}, which trains a grasp \dqn and a push \dqn then selects an action with the highest Q-value out of the two networks to execute.
    \item \mcts as described in Section~\ref{sec:mcts}. This is adapted from~\cite{huang2021visual}, but we use here a simulator to predict the next state instead of the originally used learned model, for fair comparisons. 
    \item \ppn as described in Section~\ref{sec:pushnet}. \ppn proposes push actions based on their predicted Q-values and the robot executes those actions until the target object can be grasped according to \gpn.
\end{enumerate*}


\noindent \textbf{Simulation Studies.} We ran our method and the three alternative methods on 22 cases~\cite{huang2021visual} and 10 cases~\cite{zeng2018learning, xu2021efficient}, in simulation first.
Tables~\ref{tab:sim22table} and~\ref{tab:sim10table} show the overall performance of the four methods, where \mcts based methods are limited to a budget of $50$ iterations per test case.
%
In this paper, we denote the tree search methods with different budgets of search iterations as \mcts-10/20/50 and \ours-10/20/50, where the suffix denotes the iterations limit.
%
The 22 cases are generally harder to solve than the 10 cases, where the target object can be retrieved after one push action.
The time metric records the average time (out of 5 trials) for retrieving the object, including planning and execution times.
%
%

\begin{figure}[ht!]
\vspace{1mm}
    \centering
    \includegraphics[width = .97\linewidth]{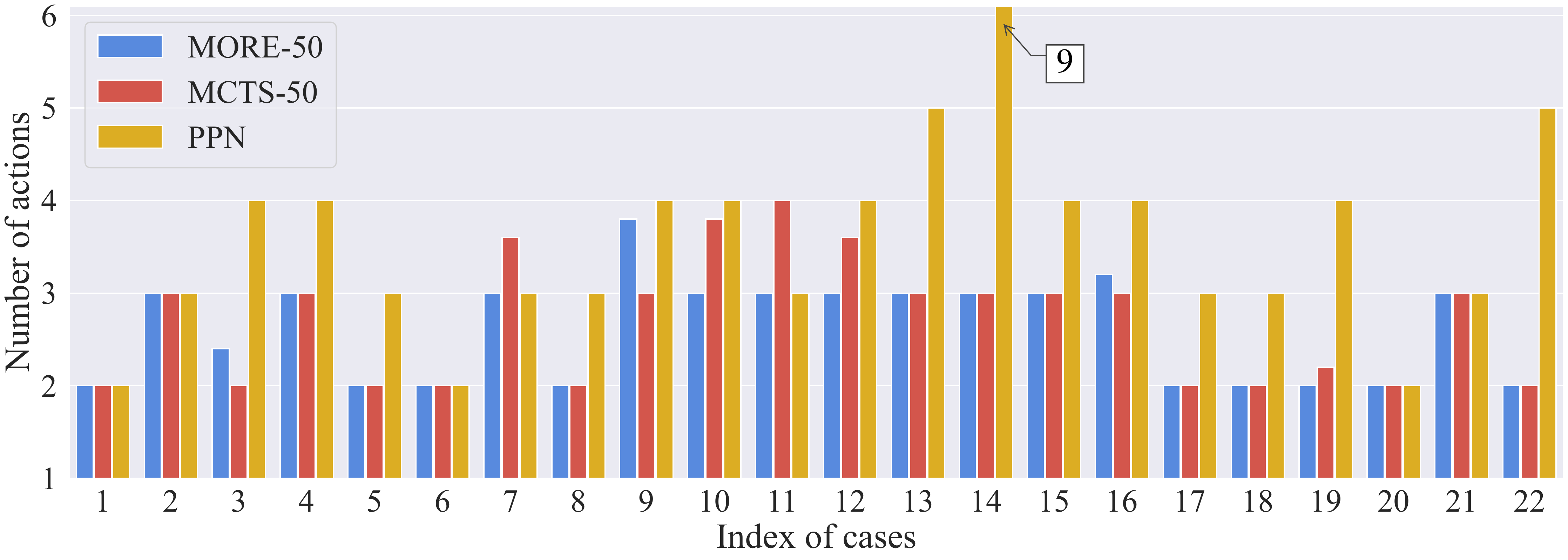}
    \caption{\label{fig:22-test-num}
        The average number (out of 5 trials) of action used to solve one case for 22 cases.
    } 
\vspace{-2mm}    
\end{figure}

\begin{figure}[ht!]
\vspace{1mm}
    \centering
    \includegraphics[width = .97\linewidth]{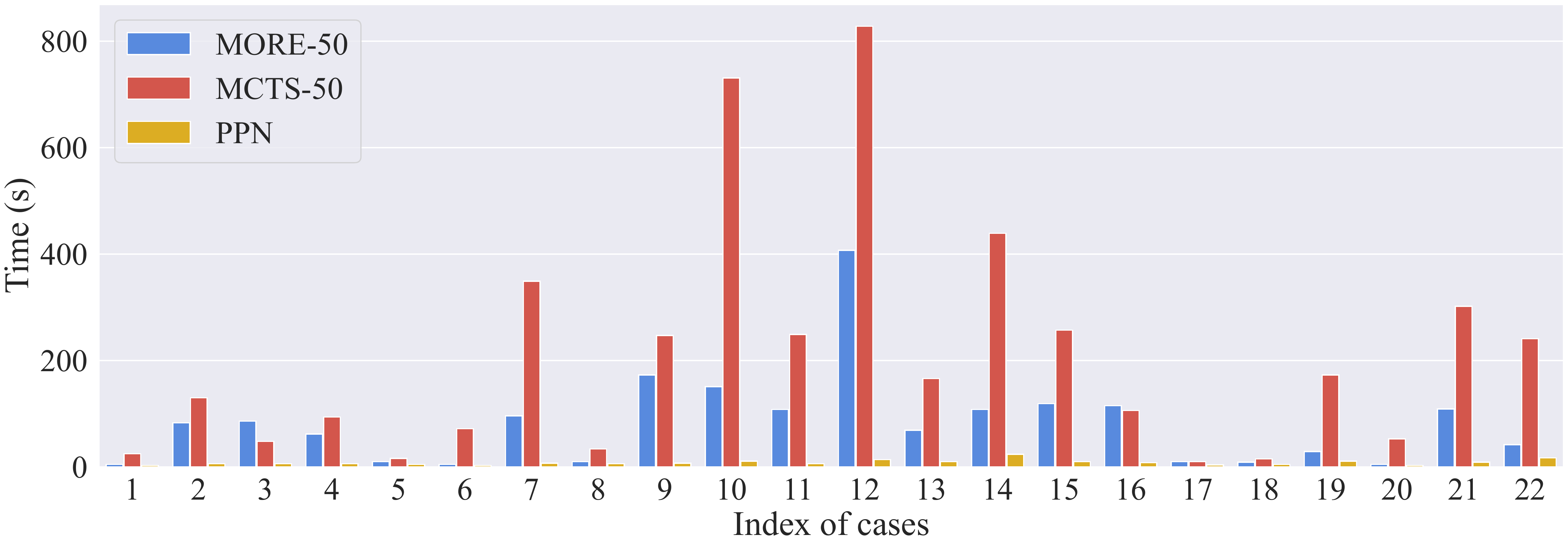}
    \caption{\label{fig:22-test-time}
        The average time (of 5 trials) used to solve one case for 22 cases.
    } 
\vspace{-1mm}    
\end{figure}

For the baseline \gopg, results on 10 cases are directly quoted from the paper (at the time of our submission, we could not obtain the trained model or the information necessary for fully reproducing \gopg).
\ours uses the fewest number of actions to solve the task.
Performance details on 22 cases can be found in Fig.~\ref{fig:22-test-num} for the number of actions and \ref{fig:22-test-time} for the running time.
\ppn is fast as it is a one-stage DNNs solution. It learned a policy that creates free spaces around the target object, but it is less consistent and less stable than the tree search solutions. From our observation, \ppn can propose non-prevailing pushing actions. 
%
\mcts provides a consistent and good quality solution, but requires a much longer planning time.
\ours, combining the benefits of both, reduces the planning time and delivers high-quality solutions.

\begin{table}[ht!]
    \centering
    \begin{tabular}{c|c|c|c|c}
        & Num. of Actions & Time & Completion & Grasp Success   \\ \hline
        \ours-50 & $\mathbf{2.61}$ & $82$s & $100\%$ & $99.2\%$ \\ \hline
        \mcts-50~\cite{huang2021visual} & $2.69$ & $208$s & $100\%$ & $99.1\%$ \\ \hline
        \ppn & $3.68$ & $8$s & $100\%$ & $97.7\%$ \\ \hline
    \end{tabular}
    \vspace{2mm}    
    \caption{Simulate experiment results for 22 cases~\cite{huang2021visual}. Budgets of \mcts and \ours are limited up to 50 iterations.}
    \label{tab:sim22table}
\vspace{-3mm}    
\end{table}

\begin{table}[ht!]
    \centering
    \begin{tabular}{c|c|c|c|c}
        & Num. of Actions & Time & Completion & Grasp Success   \\ \hline
        \ours-50 & $\mathbf{2.10}$ & $16$s & $100\%$ & $100\%$ \\ \hline
        \mcts-50~\cite{huang2021visual} & $2.20$ & $32$s & $100\%$ & $93.4\%$ \\ \hline
        \ppn & $2.70$ & $4$s & $100\%$ & $95.0\%$ \\ \hline
        \gopg~\cite{xu2021efficient} & $2.77$ & $-$ & $99.0\%$ & $90.0\%$ \\ \hline
    \end{tabular}
    \vspace{2mm}    
\caption{Simulate experiment results for 10 cases~\cite{xu2021efficient}. Budgets of \mcts and \ours are limited up to 50 iterations.}
    \label{tab:sim10table}
\vspace{-2mm}    
\end{table}

\begin{figure}[ht!]
    \centering
    \includegraphics[width = \linewidth]{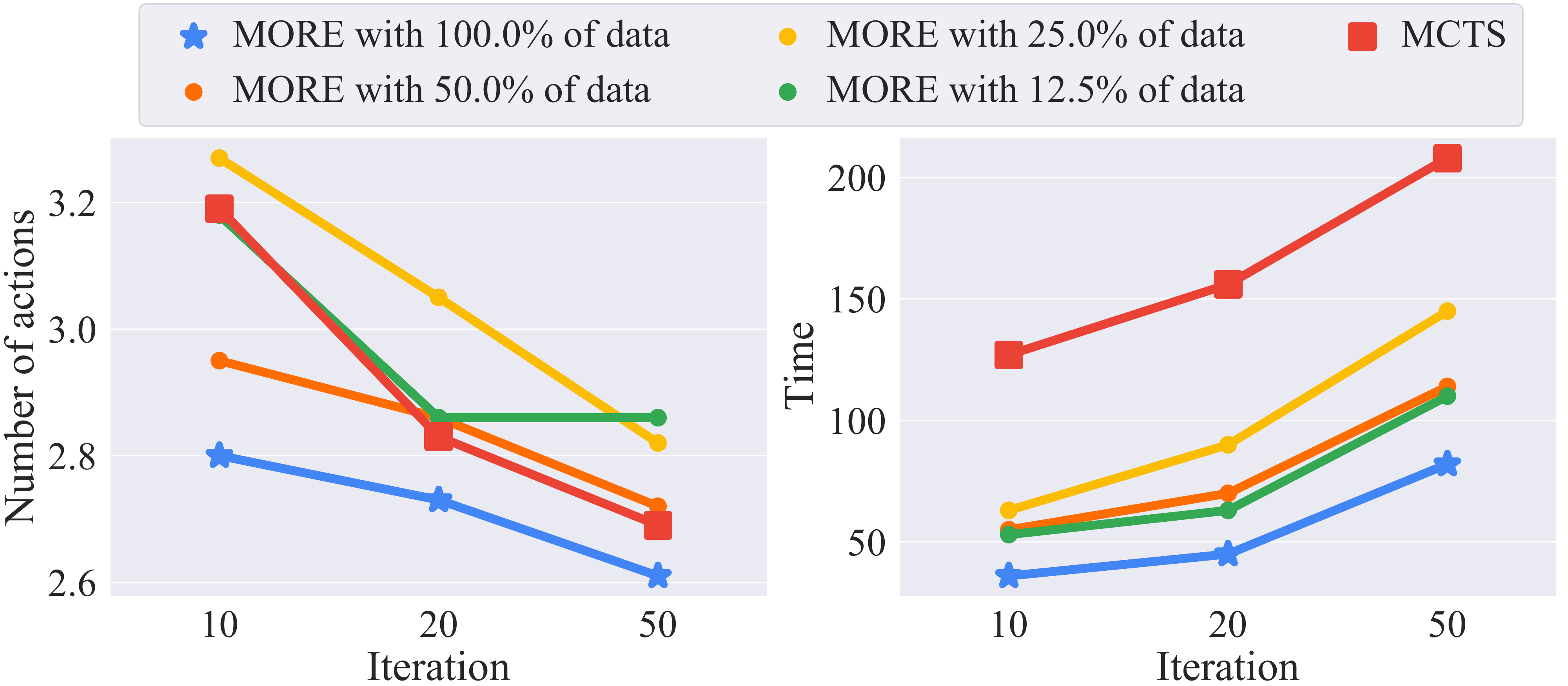}
    \caption{\label{fig:bar-num-time}
        Different amounts of training data are used to train \ppn, which are evaluated on \ours with different budgets (iteration).
        This is the evaluation of the 22 cases.
    } 
\vspace{-2mm}    
\end{figure}

\noindent \textbf{Ablation Studies}
Although the data generated by MCTS for training \ppn is free because it is collected fully automatically in simulation, we set to explore data efficiency in training, which can be important for building larger models in practice.
For this purpose, we collected $243$ training cases ($65384$ transitions in 30 hours with PyBullet) with \mcts as described in Section~\ref{sec:mcts}.
Training on \ppn on all data used around 22 hours.
As shown in Fig.~\ref{fig:bar-num-time}, we tested \mcts and \ours with different budgets. Also, \ours is trained on different numbers of training data.
Clearly, the problem can be solved by all tested methods with fewer actions when the search iteration limits are increased.
But the time for solving the problem also increases as a consequence.
The proposed \ours technique can retrieve target objects with only $2.8$ executed actions and using only $10$ iterations of MCTS that last $36$ seconds on average. This is close to the best that MCTS without \ppn can achieve, $2.69$ actions, after $50$ iterations that last $208$ seconds. When we limit the number of iterations of MCTS (without \ours) to $10$, the number of executed actions increases to $3.19$, and the search time remains relatively high ($127$ seconds). This clearly shows the out-performance of the proposed approach in terms of both time and action efficiency. 

%
%

\subsection{Robot Experiments}
\begin{figure}[ht!]
    \centering
    \includegraphics[width = \linewidth]{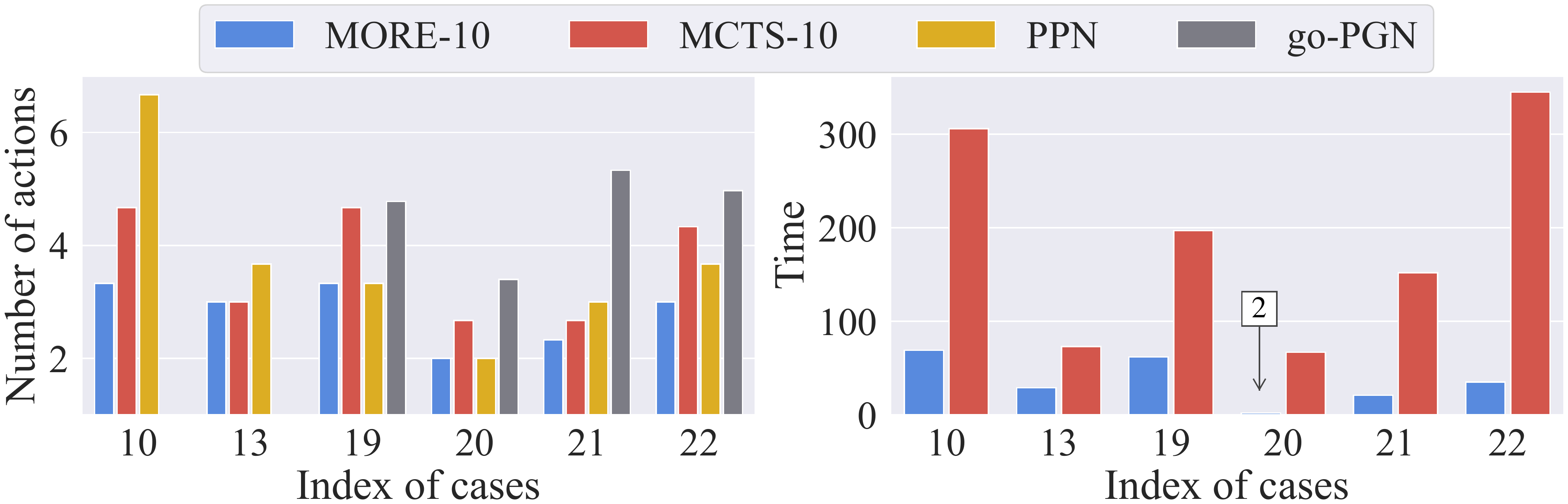}
    \caption{\label{fig:real-num-time}
        The number of action and time used on solving six cases.
        The budget is up to $10$ iterations for \mcts and \ours.
    } 
\vspace{-2mm}    
\end{figure}

We evaluated the four methods on six real test cases (four from~\cite{xu2021efficient} and two from~\cite{huang2021visual}).
These six test cases are representative in that they contain more objects and often require at least two push actions to solve.
For these real experiments, the results are shown in Table.~\ref{tab:real6table} and Fig.~\ref{fig:real-num-time}. 
The budget of \mcts and \ours is limited to $10$ iterations. We note that the results for \gopg are taken from~\cite{xu2021efficient}. The execution time of \ppn is not listed in Table.~\ref{tab:real6table} as it is a near-constant small value as we had in the simulation experiments. 
%
From the result, we observe only negligible performance degradation in comparison to simulation, which may be due to differences in friction, slight differences in the dimensions of the objects between simulation and real world, statistical error, or a combination of these.
Overall, the sim-to-real transfer was very successful and showed that \ours can learn in simulation and directly apply the learned skill to real-world tasks.
We assume models of objects are known, such that simple pose estimation can be used to locate objects in the real world and placed in simulation for planning. We could also use sophisticated tracking systems~\cite{wen2020se, wen2021bundletrack, mitash2020scene} for general purpose.

\begin{figure}[ht!]
    \centering
    \includegraphics[width = 0.15 \linewidth, trim = {0, 0, 0, 0}, clip]{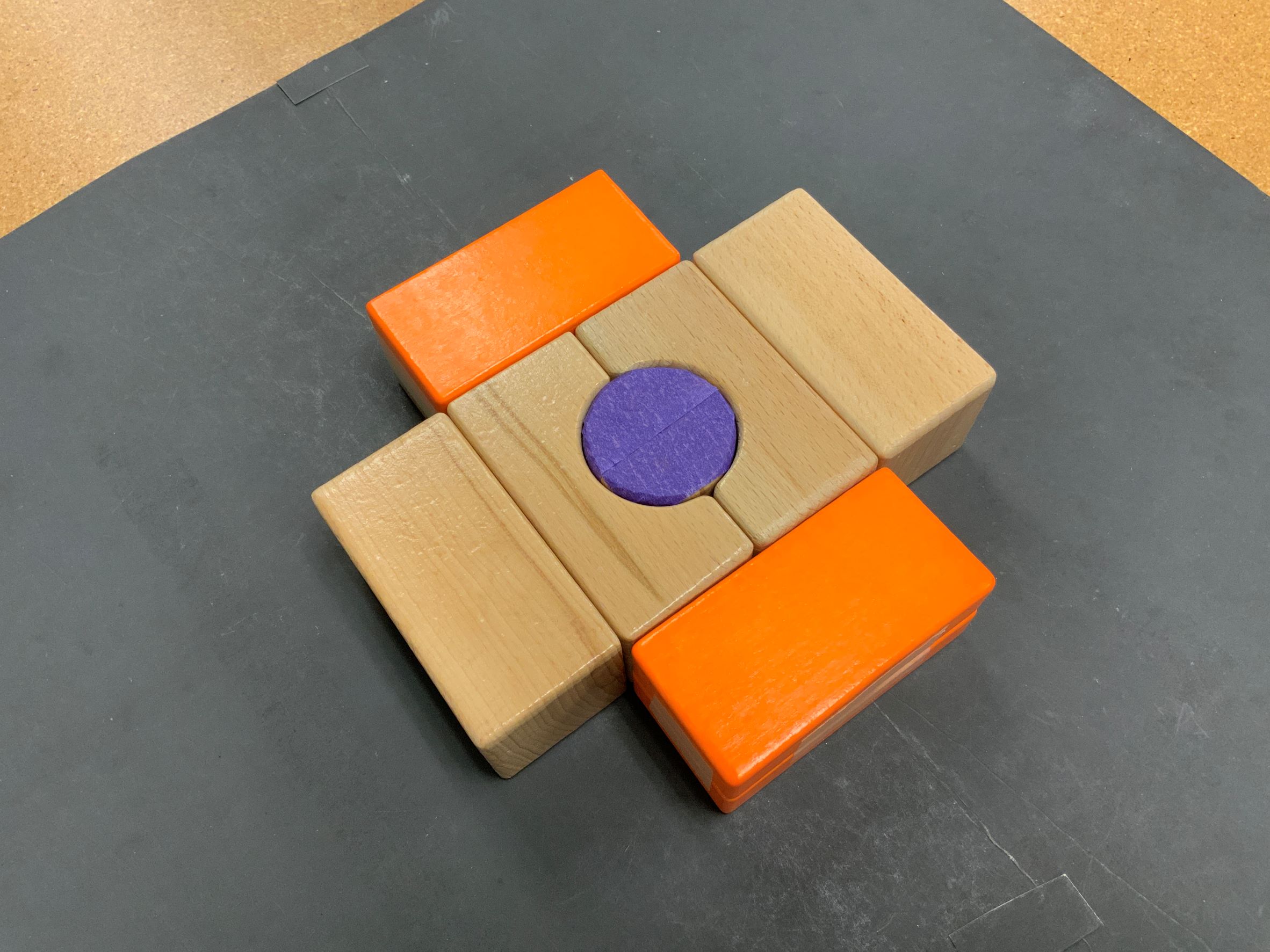} 
    \includegraphics[width = 0.15 \linewidth, trim = {0, 0, 0, 0}, clip]{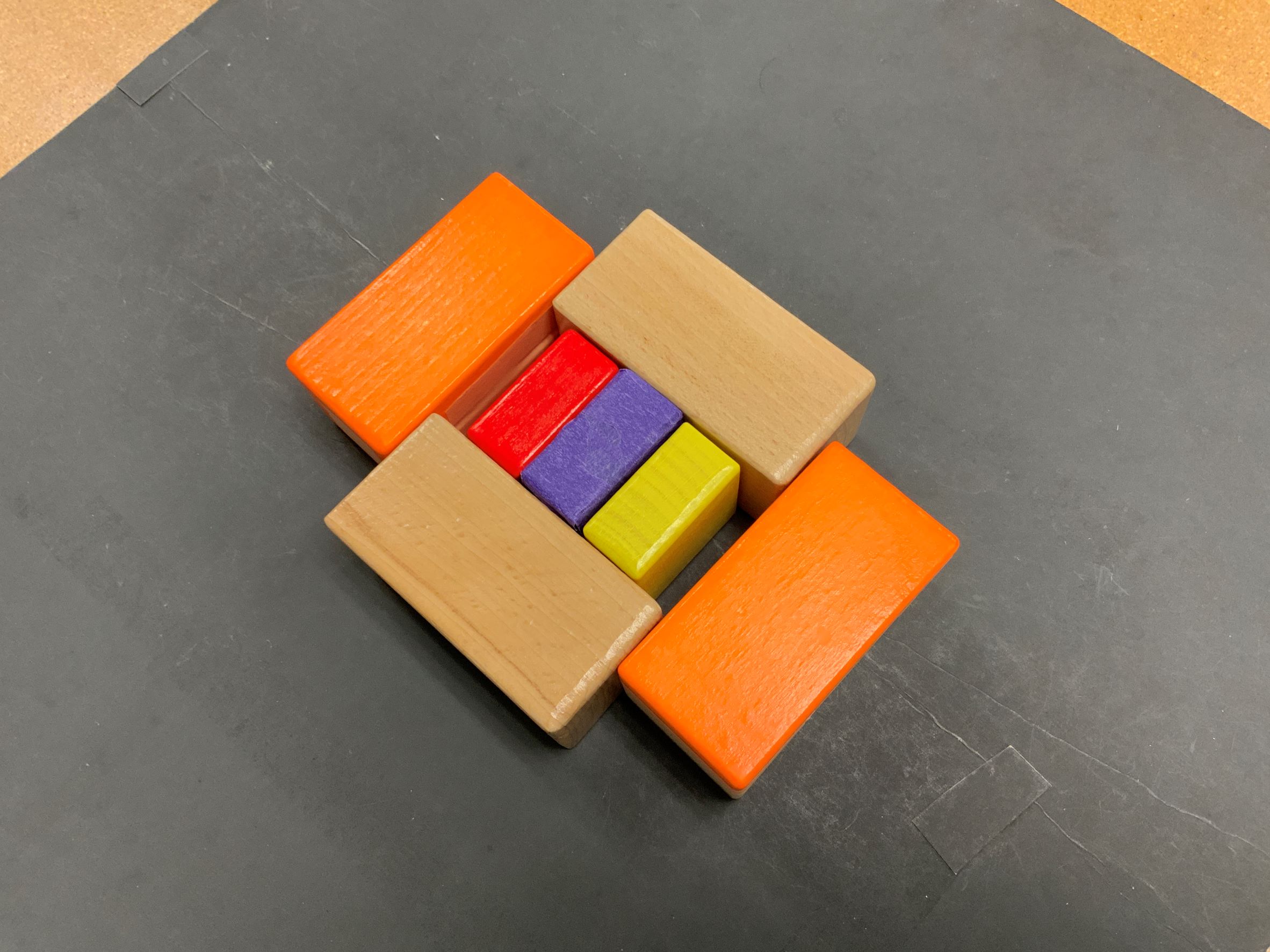} 
    \includegraphics[width = 0.15 \linewidth, trim = {0, 0, 0, 0}, clip]{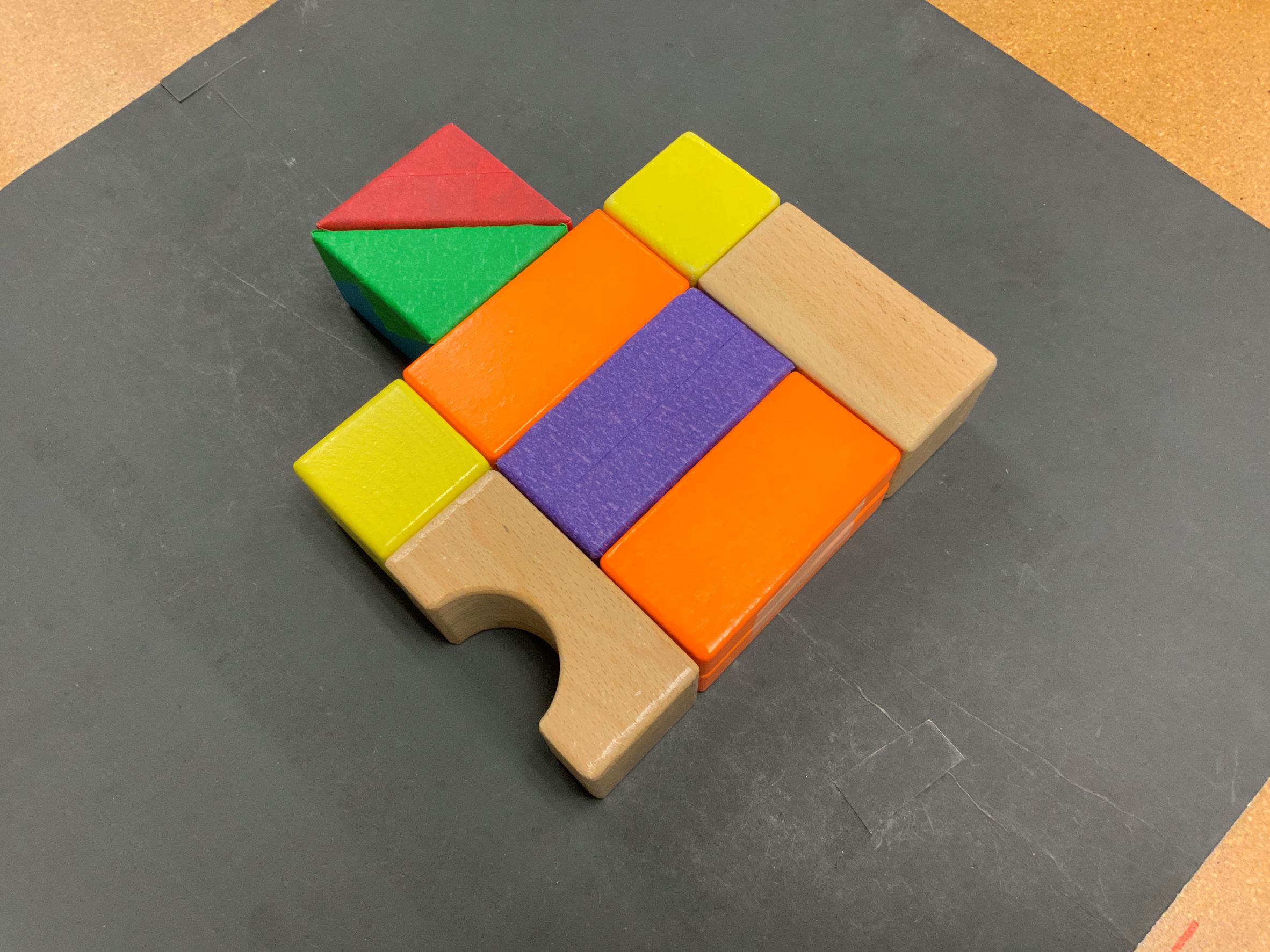} 
    \hspace{0mm}
    \includegraphics[width = 0.15 \linewidth, trim = {0, 0, 0, 0}, clip]{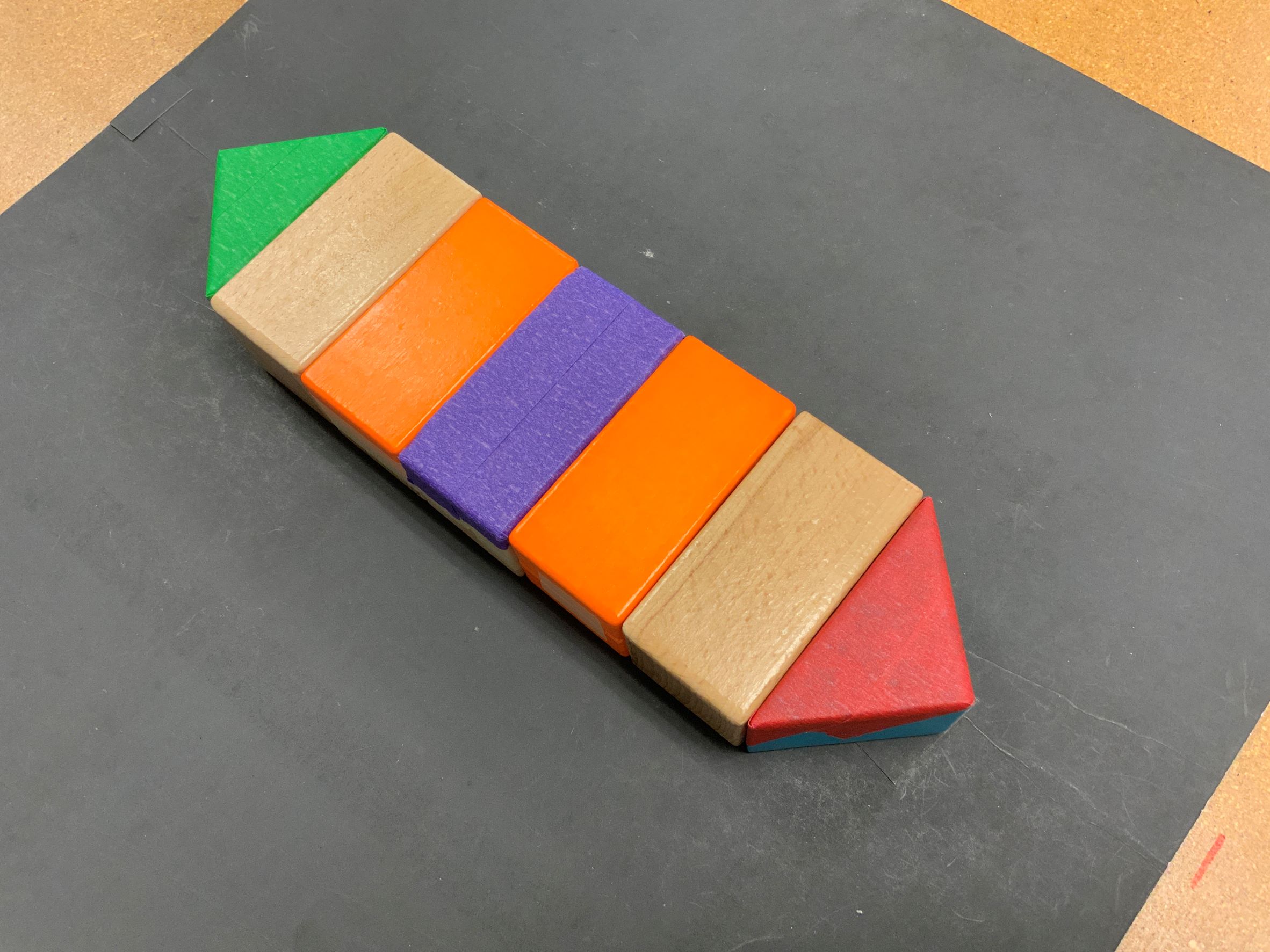} 
    \includegraphics[width = 0.15 \linewidth, trim = {0, 0, 0, 0}, clip]{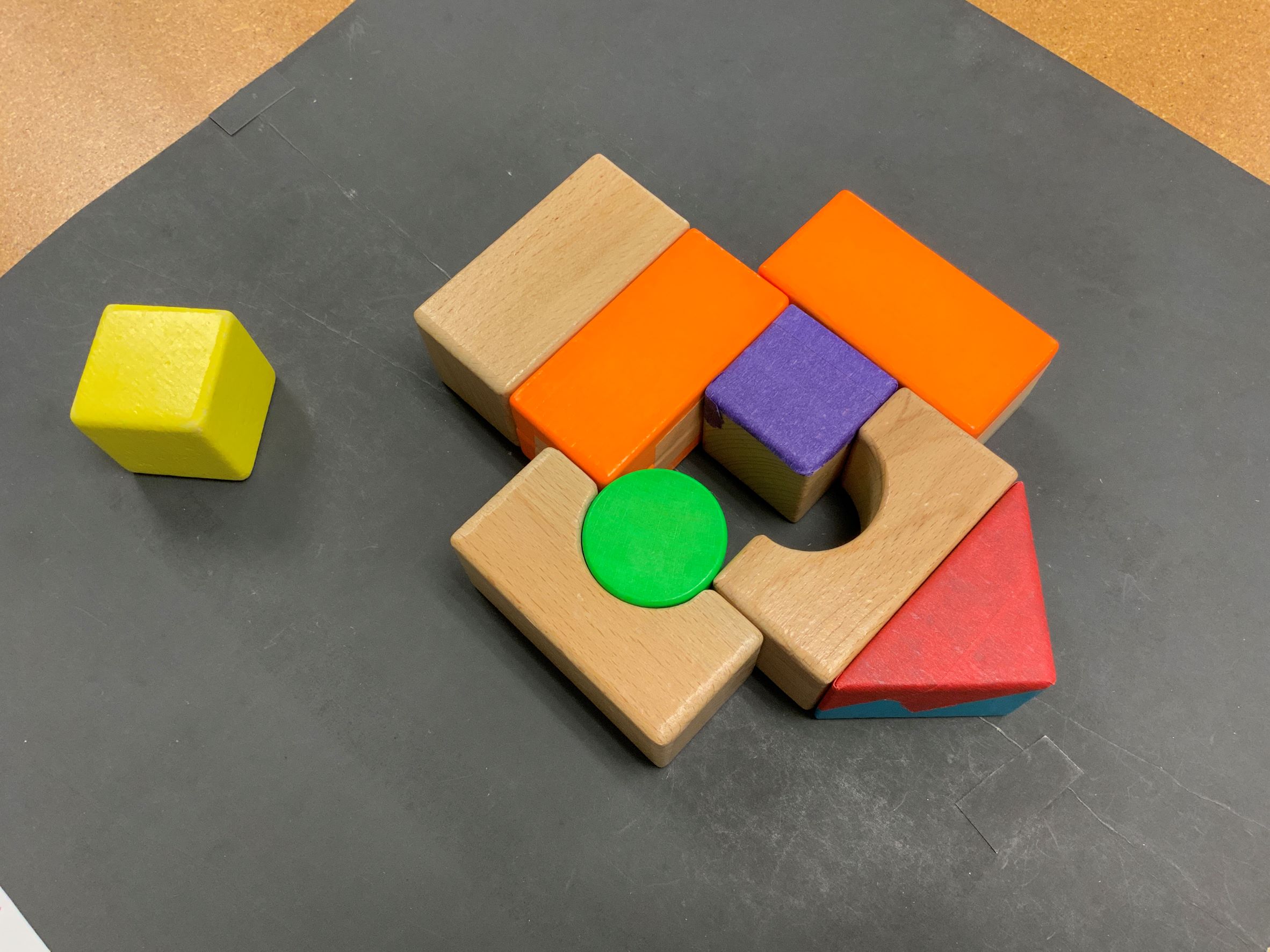} 
    \includegraphics[width = 0.15 \linewidth, trim = {0, 0, 0, 0}, clip]{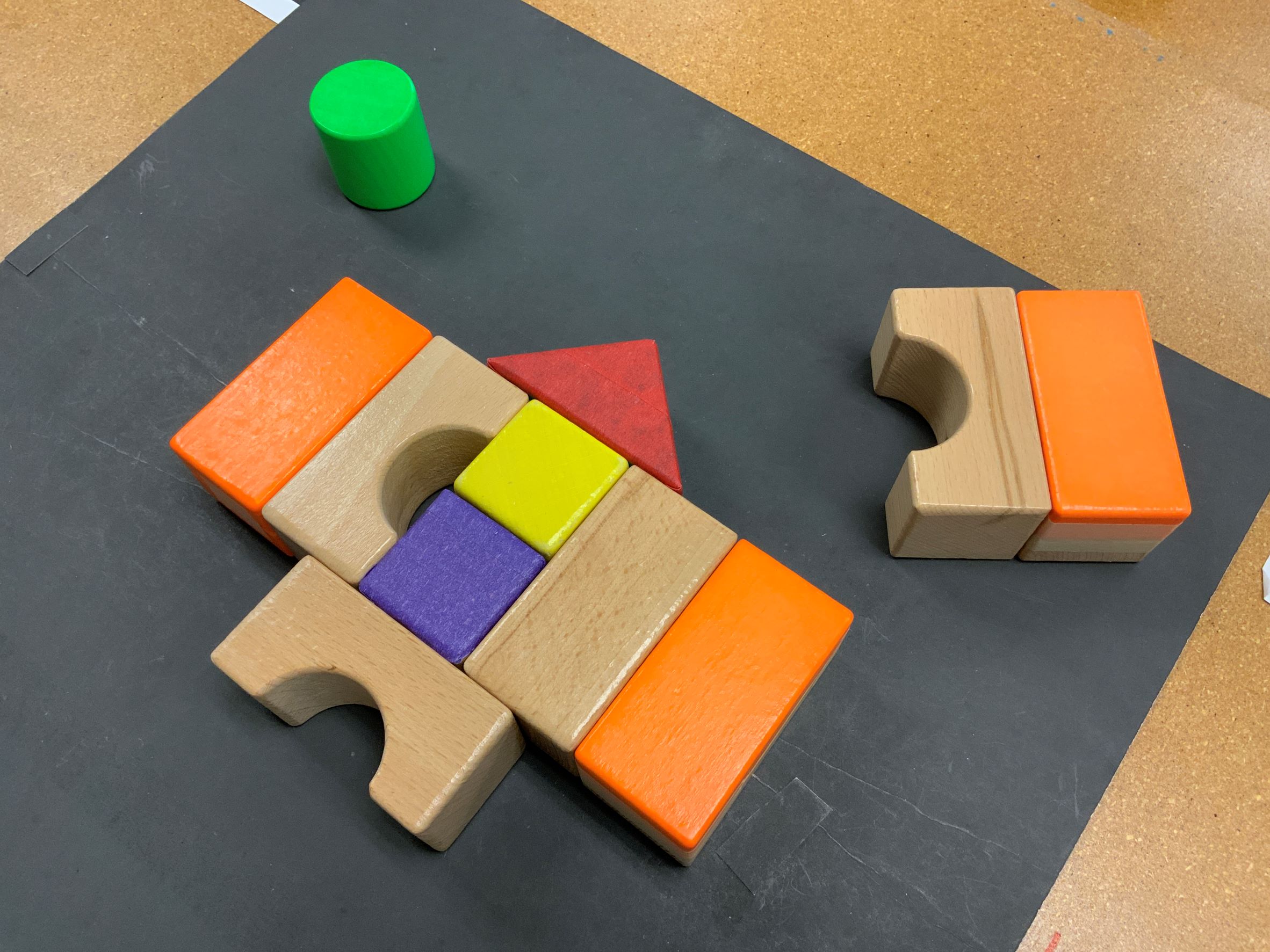}
    \vspace{1mm}
    \caption{\label{fig:real-cases}
    Manually generated cases similar to~\cite{huang2021visual, xu2021efficient}. The target object is masked in \textcolor{RoyalPurple}{purple}. These cases are used also in simulation experiments as shown in Fig.~\ref{fig:testcases}.
    }
    \vspace{-4mm}
\end{figure}

\begin{table}[ht!]
    \centering
    \begin{tabular}{c|c|c|c|c}
        & Num. of Actions & Time & Completion & Grasp Success   \\ \hline
        \ours-10 & $\mathbf{2.83}$ & $36s$ & $100\%$ & $100\%$ \\ \hline
        \mcts-10~\cite{huang2021visual} & $3.67$ & $190s$ & $100\%$ & $95.8\%$ \\ \hline
        \ppn & $3.72$ & $3s$ & $94.5\%$ & $95.8\%$ \\ \hline
        \gopg~\cite{xu2021efficient} & $4.62$ & $-$ & $95.0\%$ & $86.6\%$ \\ \hline
    \end{tabular}
\vspace{2mm}    
    \caption{Real experiment results for six cases as shown in Fig.~\ref{fig:real-cases}. The budget of \mcts and \ours is limited to 10 iterations.
    For \gopg, only the first four cases apply, and results are from~\cite{xu2021efficient}. 
    Only planning time is recorded (robot execution was intentionally slowed down for safety).
    The computation time for \ppn to solve a task is 3 seconds on average (estimated).
    }
    \label{tab:real6table}
\vspace{-2mm}    
\end{table}

\section{Discussion and Conclusion}\label{sec:conclusion}
The main limitation of this work is that we need to know the models of the objects to do the planning. One possible solution is instead of using an explicit simulator, we can use a learned model~\cite{huang2020dipn} to simulate the push results. Generalization to novel objects could then be possible.
We can further utilize the Push Prediction Network to estimate the simulation (rollout) result instead of using a physics engine. However, this can introduce additional uncertainties that typically result from using DNNs, which can cause unexpected behaviors such as pushing objects out of the workspace. 
Building on the know-hows gains from developing \ours, we are exploring other real-world robotic manipulation tasks that would benefit from the S2$\to$S1 search-and-learn philosophy. 
We point out that \ours can be further sped up by implementing a parallel version of \mcts, as we only utilized a single CPU thread in our implementation and \ppn (on GPU) is not being used most of the time. 


\bibliographystyle{IEEETran}
\bibliography{references}

\end{document}

%% file: related-works.tex
\noindent {\bf Grasping.} Grasping approaches can be classified as being \emph{analytical} or \emph{data-driven}~\cite{10.1109/TRO.2013.2289018}. 
Analytical methods examine precise object models to predict the stability of a grasp based on {\it force-closure} or {\it form-closure}~\cite{grasping,liang2019pointnetgpd,doi:10.1177/0278364912442972}.
However, high precision 3D models of objects, e.g., YCB objects~\cite{calli2017yale}, are hard to come by. 
In addition, other material properties, such as friction and inertia, are challenging to measure.
These challenges have given rise to data-driven methods that learn from data, where many works focus on isolated objects~\cite{DBLP:conf/iros/BoulariasKP11,DBLP:conf/iccv/MousavianEF19,gabellieri2020grasp,lu2020multifingered}.
Recently, grasping in clutter has received more attention~\cite{DBLP:conf/aaai/BoulariasBS14, mahler2017dexnet,kalashnikov2018qtopt, fang2020graspnet, wen2021catgrasp}.
Convolutional Neural Networks are widely used to construct grasp proposal networks such as Dex-net 4.0~\cite{mahler2019learning}, which are trained to detect 6D grasp poses in point clouds~\cite{DBLP:journals/corr/PasGSP17}.
%
%
%
In this paper, we use a self-supervised Deep Q-Network similar to~\cite{zeng2018learning} for grasping in clutter.

\noindent {\bf Singulation.} 
Singulation, i.e., isolating specific object(s) from the rest~\cite{6224575}, is necessary for object retrieval.
%
%
Usually, a sequence of pushing and grasping actions is used to clear the clutter that surrounds the target object.
In~\cite{10.1007/978-3-030-28619-4_32}, a \emph{model-free} method was used to learn a reactive pushing policy without long-horizon reasoning.
Later, other model-free reinforcement learning algorithms~\cite{zeng2018learning, 8560406} used learned push policies to improve grasping.
%
%
In contrast to existing work on singulation, we explicitly seek to minimize the number of actions needed to isolate a target object for grasping sufficiently. 
%

\noindent {\bf Object Retrieval.} 
Object retrieval from clutter, the focus of this study, can be viewed as a form of rearrangement planning \cite{DBLP:journals/corr/abs-1912-07024,GaoFenYuRSS21,YuRSS21}.
%
%
%
Online planning for object search with partial observations has been discussed in~\cite{8793494}.
Retrieving objects under occlusion was also recently considered in~\cite{DBLP:journals/corr/abs-1903-01588} where parallel-jaw and suction grasping were used along with pushing to de-clutter surroundings of target objects. 
%
%
%
A model-free reinforcement learning technique has also been used for searching for objects in~\cite{DBLP:journals/corr/abs-1911-07482}.
In~\cite{kurenkov2020visuomotor},  an agent was trained to find a continuous trajectory of a gripper that pushes away clutter or pushes the target object to free space, mimicking human-like behavior.
%
%
A human in-the-loop solution was proposed in~\cite{DBLP:journals/corr/abs-1904-03748} help with searching for objects in clutter.
%
%
A deep Q-Learning method~\cite{xu2021efficient} considers a similar task and setup but uses additional primitives such as sliding objects from the top.
%
%
%
%
%
Our work partially builds on~\cite{huang2021visual}, which explores the use of MCTS for the same object retrieval problem. 
%
In contrast to existing object retrieval works, we focus on developing the machinery that enables the S2$\to$S1 philosophy to reduce the computational burden of the related search problem while using real robots and objects.